\documentclass[letterpaper, 10 pt, conference]{ieeeconf}  

\IEEEoverridecommandlockouts                              

\usepackage{cite}
\usepackage{amsmath,amssymb,amsfonts}
\usepackage{algorithmic}
\usepackage{graphicx}
\usepackage{textcomp}
\usepackage{xcolor}
\usepackage{stfloats}
\def\BibTeX{{\rm B\kern-.05em{\sc i\kern-.025em b}\kern-.08em
    T\kern-.1667em\lower.7ex\hbox{E}\kern-.125emX}}
\usepackage{etoolbox}
\newcommand{\norm}[1]{\left\lVert#1\right\rVert}

\title{\LARGE \bf
Classifying Bicycle Infrastructure Using On-Bike Street-Level Images
}

\author{Kal Backman, Ben Beck and Dana Kuli\'c
\thanks{ This project was funded by an Australian Government Department of Infrastructure, Transport, Regional Development and Communications `Road Safety Innovation Fund' grant. Ben Beck was supported by an Australian Research Council Future Fellowship (FT210100183). Dana Kuli\'c was supported by an Australian Research Council Future Fellowship (FT200100761). Authors are from the Department of Electrical and Computer Systems Engineering and School of Public Health and Preventive Medicine Monash University. [Kal.Backman, Ben.Beck, Dana.Kulic]@monash.edu}
}
\begin{document}

\maketitle
\thispagestyle{empty}
\pagestyle{empty}

\begin{abstract}
While cycling offers an attractive option for sustainable transportation, many potential cyclists are discouraged from taking up cycling due to the lack of suitable and safe infrastructure. Efficiently mapping cycling infrastructure across entire cities is necessary to advance our understanding of how to provide connected networks of high-quality infrastructure. Therefore we propose a system capable of classifying available cycling infrastructure from on-bike smartphone camera data.
The system receives an image sequence as input, temporally analyzing the sequence to account for sparsity of signage. The model outputs cycling infrastructure class labels defined by a hierarchical classification system. Data is collected via participant cyclists covering 7,006Km across the Greater Melbourne region that is automatically labeled via a GPS and OpenStreetMap database matching algorithm. 
The proposed model achieved an accuracy of 95.38\%, an increase in performance of 7.55\% compared to the non-temporal model. The model demonstrated robustness to extreme absence of image features where the model lost only 6.6\% in accuracy after 90\% of images being replaced with blank images.
This work is the first to classify cycling infrastructure using only street-level imagery collected from bike-mounted mobile phone cameras, while demonstrating robustness to feature sparsity via long temporal sequence analysis.
\end{abstract}

\section{Introduction}
With growing interest in sustainable transportation solutions \cite{Zhao2020, Anton2023}, cycling is at the forefront due to its high spatial penetration and efficiency over short distances \cite{Zhao2020}, low environmental impact \cite{Comi2021} and providing auxiliary health benefits \cite{Teschke2012, Mueller2015}. However, in many cities around the world, bicycle participation is low \cite{Goel2022}. In Australia, less than 2\% of trips are by bike, and 60\% of people report how unsafe they feel while riding as the number one barrier to riding a bike \cite{Pearson2023, Pearson2023_2}. Significant increases in bike riding could be realised should connected networks of high-quality bike infrastructure be provided that meets the needs of people of all ages and abilities \cite{Pearson2022}. Specifically, the provision of infrastructure that separates people from motor vehicles (e.g. protected bike lanes or off-road paths) is critical to overcoming negative user experiences \cite{Winters2011} and can dramatically enhance safety and increase cycling \cite{Molenberg2019, Aldred2019, Kraus2021}.

In recognition of the many benefits of getting more people riding as part of everyday life, governments globally are investing in bicycle infrastructure with the aim of increasing the safety and uptake of bike riding. In order to optimise the existing network and plan for new infrastructure, governments need reliable city-wide bike infrastructure spatial databases. However, bicycle network data from governments or crowd-sourced projects often suffer from inconsistencies and poor quality, hampering efforts to enhance monitoring, evaluation and investment in the bike network \cite{Viero2024}. With the growing capability of on-bike video technology from dedicated cameras and mounted smartphones, such infrastructure characterisation could be carried out using crowd-sourced video data. When coupled with autonomous processing of camera imagery using deep learning classification algorithms, practical large-scale continual monitoring and reporting becomes achievable. 

In this work we take the initial steps to accomplish this goal by developing and evaluating a pipeline capable of classifying cycling infrastructure from on-board street-level imagery. The pipeline forms the backbone of the crowd sourced cycling infrastructure mapping and route recommendation system by transforming video streams synchronized with GPS coordinates into a set of classified coordinates that indicate the cycling infrastructure present at a given location.

\subsection{Related work}
Prior works related to deep learning classification for cycling infrastructure are limited. \cite{Armin2024} demonstrated the use of 360 degree LiDAR scans to produce point clouds that are segmented into points that are able to be ridden on. However the approach was limited in practicality due to the large costs associated with the 5 LiDARs used, whilst also not providing sufficient detail about the type of cycling infrastructure due to classifying all ridable points as roads.  

\cite{Ito2021} assigned bikeability scores to regions across Singapore and Tokyo based on 34 indicators which included information extracted from street view imagery. Information extracted from street view imagery includes segmenting the ratio of visually aesthetic regions within the image and applying object detection of vehicles to estimate traffic volume. The main limitation of \cite{Ito2021} was that it operated on temporally static imagery and did not aim to detect any forms of cycling infrastructure.

\begin{figure*}[b]
\centering
\includegraphics[width=1.\textwidth]{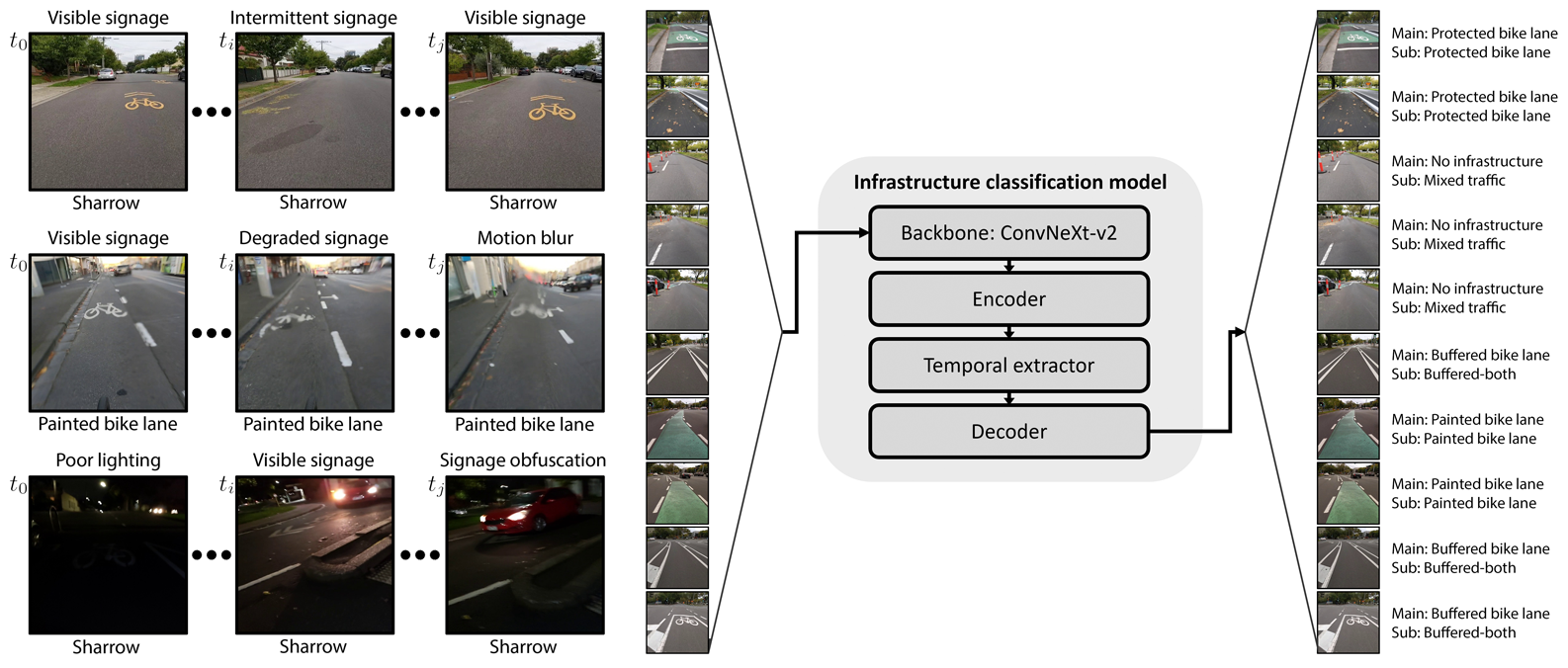}
\caption{Overview of the infrastructure classification model. (Left) Estimating cycling infrastructure from a single image is difficult due to the temporal sparsity and occlusion of signage. (Right) Summary of the model’s architecture which consists of an image sequence input from which image features are extracted and compressed onto a latent vector.}
\label{Overview}
\end{figure*}

\cite{Knura2021} aimed to pool crowd-sourced imagery from social media for use in urban planning of cycling infrastructure. The approach deployed an object detection model to detect and classify bicycles as either stationary or moving for use in the planning of bicycle parking infrastructure availability. The approach is limited since it only investigates parking, rather than aiming to classify infrastructure.

Alternative forms of infrastructure detection and classification exist outside of the cycling domain and include road infrastructure segmentation to highlight signage, guard rails, roads \cite{Chen2019, Gao2017} and speed bumps \cite{Arun2020}, used in the perception of autonomous vehicles. The condition of such road infrastructure can be further determined through object detection algorithms to highlight cracks and signage degradation \cite{Maeda2018} or by classifying road images by their afflicted environmental conditions \cite{Dewangan2021, Sirirattanapol2019}. Infrastructure detection can also be carried out with aerial imagery, using drones to detect road \cite{Jiang2023} and power line \cite{Varghese2017, Nguyen2019} infrastructure, or satellite imagery to detect utilities \cite{Oshri2018}, transportation \cite{Henry2018} and building \cite{Francisco2021, Robinson2022} infrastructure. 

\subsection{Model overview}
In this paper, we propose to classify cycling infrastructure from on-bike camera data. An overview of the model architecture can be found in Figure~\ref{Overview}. Single image infrastructure classification is difficult due to the degradation and occlusion of relevant signage, environmental conditions and the temporal sparsity of relevant information from intermittent signage as seen in Figure.~\ref{Overview} (left). Furthermore, cycling infrastructure can be difficult to estimate due to subtle differences in features and the ambiguity in determining when one infrastructure class starts and the other stops. Therefore a hierarchical infrastructure classification model is proposed that temporally analyses image sequences to predict cycling infrastructure classes.

The cycling infrastructure classification model takes an image sequence as input. Each image is passed through the ConvNeXt-V2 \cite{ConvNext} backbone to produce a feature embedding that is subsequently compressed onto a one dimensional latent vector via an encoder. The latent vectors are temporally analyzed using the Temporal extractor which employs self-attention \cite{Attention} to compare latent vectors across time to one another. The temporally analyzed latent vectors are then decoded into a hierarchical classification scheme to estimate the main and sub classes of each of the input images.   

The main contributions of the proposed work are as follows:
\begin{enumerate}  
\item The first published work to classify cycling infrastructure using street-level imagery;
\item A proposed image classification model robust to absent image features by analyzing long temporal sequences; and 
\item Large scale participant gathered dataset of street-level imagery attained from on-bike video recordings.    
\end{enumerate}

\section{Infrastructure classification model}\label{ModelSection}
\begin{figure}[b!]
\centering
\includegraphics[width=1.0\columnwidth]{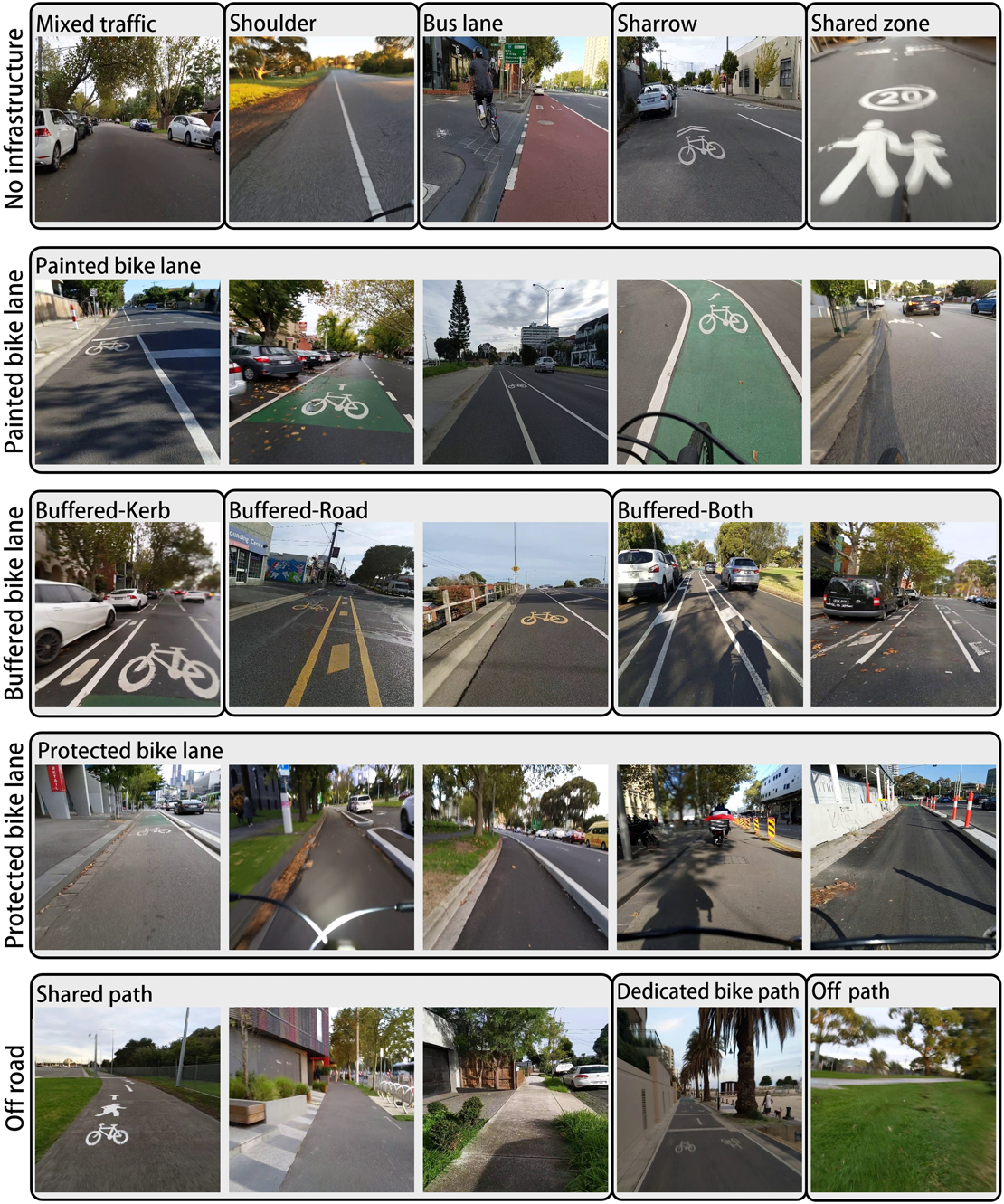}
\caption{Example images of the different cycling infrastructure classes and their associated groupings into main and sub classes.}
\label{ExampleClasses}
\end{figure}

To classify video frames into specific cycling infrastructure classes, a hierarchical classification model is proposed that takes in a sequence of video frames as input and outputs one of five main classes and one of thirteen sub-classes. The five main classes and their associated sub-classes are derived from \cite{Debjit} and are:
\begin{enumerate}
    \item \textbf{No infrastructure}: mixed traffic, shoulder, bus lane, shared zone, sharrow
    \item \textbf{Painted bike lane}: painted bike lane
    \item \textbf{Buffered bike lane}: buffered (kerb-side), buffered (road-side), buffered (both)
    \item \textbf{Protected bike lane}: protected bike lane
    \item \textbf{Off road}: shared off-road path, dedicated off-road bike path, off path
\end{enumerate}
Example images for each of the classes can be seen in Figure.~ \ref{ExampleClasses}.

\vspace{1.5cm}
The infrastructure classification model's architecture consists of 4 core components:
\begin{enumerate}
    \item Backbone
    \item Encoder
    \item Temporal extractor
    \item Decoder
\end{enumerate} 
\(N\) images are sampled from a video sequence with a time spacing of \(t\)-seconds apart and image resolution of \(512 \times 512 \times 3\) which is down sampled from the raw \(1280\text{w} \times 960\text{h} \times 3\) input video. Each image in a video sequence is fed through the backbone network whose objective is to extract distinguishing features from input images using the ConvNeXt-V2-nano \cite{ConvNext} architecture to produce feature representations of resolution \(16 \times 16 \times 640\). 

The feature representations are subsequently fed into the encoder to produce latent vectors of length \(\mu\). The objective of the encoder is to compress the features produced by the backbone into a single vector such that they can be analyzed by the temporal extractor.

The latent vectors representing the \(N\) sequential images are then analyzed temporally using the temporal extractor which adds sinusoidal positional encodings then subsequently deploys 3 sequential multi-headed attention blocks to produce latent vectors of length \(\mu\) for each input image \cite{Attention}. The objective of the temporal extractor is to compare each of the latent vectors of the encoder to one-another across time.  By paying attention across time, the classifier is robust to frames with insufficient information within a single image, e.g. due to the lack of continual signage / road markings. 

The temporally analyzed latent vectors are then passed through the decoder which produces logits for each of the five main classes and each of the thirteen sub-classes. A softmax activation function is applied to the five main classes to produce probabilities for each of the main classes where the sub-classes associated with the most probable main class have a softmax activation function applied to them to produce probabilities for the relevant sub-classes. 

The loss function comprises of two cross-entropy losses between the predicted main-class \(\hat{y}_m\) and the true main-class label \(y_m\), and between the predicted sub-class  \(\hat{y}_s\) and the true sub-class label \(y_s\).
Each class label contains a confidence score indicating the confidence in the accuracy of the ground truth label attained from Section \ref{LabelingSection}, which is used to weight the total loss. The total loss used to train the model is: 
\begin{equation}
\text{Loss} = \text{Conf} * (w_m * \text{CE}(\hat{y}_m, y_m) + w_s * \text{CE}(\hat{y}_s, y_s)). 
\end{equation}
Where \(w_m\) and \(w_s\) are scaling weights for the main-class loss and the sub-class loss respectively and are given a value of 1.0 and 0.5.

\subsection{Model training}
The infrastructure classification model is trained in a two-step process:
\begin{enumerate}
 \item Backbone training
 \item Temporal training
\end{enumerate}
\subsubsection{Backbone training}\label{BackBoneTrainingSect}
The objective of the initial training phase is to speed up the temporal training process by pretraining the backbone network so that it can extract useful image features for temporal analysis in the subsequent training step. The advantage of backbone pretraining is to allow for larger batch sizes due to lower memory requirements and quicker optimization processing times due to only processing a single image within a batch instead of the full sequence of \(N\) images. 

During backbone training, the model consists of the backbone, encoder and decoder, omitting the temporal extractor. The encoder produces latent vectors of length \(\mu = 512\) which is directly fed into the decoder. Images are sampled using a batch size of 32, where each image within the batch has augmentations applied to them. The model is trained for a total of 6 epochs using a single NVIDIA RTX 3090 during the backbone training phase. After the completion of backbone training, the hidden state representations of the backbone’s output are computed for the entire training dataset and saved to disk.

\subsubsection{Temporal training}
The objective of the temporal training phase is to teach the network how to process information across time to produce temporally consistent outputs that are robust to intermittent indications of the correct infrastructure class such as road signage.    
During temporal training the model consists of the encoder, temporal extractor and the decoder, where the backbone is omitted and instead the encoder directly receives the pre-computed feature representations from the backbone training phase as input. The advantage of learning from precomputed features is the significant reduction in training time as the majority of computational resources during the forward pass are consumed during the computation of feature representations of the backbone network. 

During the temporal training phase image sequences of length \(N=50\) with a time spacing of \(t = 1\) second are sampled using a batch size of 5 image sequences. For temporal training, distributed training is performed across two NVIDIA RTX 3090s, resulting in an effective batch size of 10. The model is trained for a further 1 epoch during the temporal training phase.

\section{Data gathering}
\begin{figure}
\centering
\includegraphics[width=1.0\columnwidth]{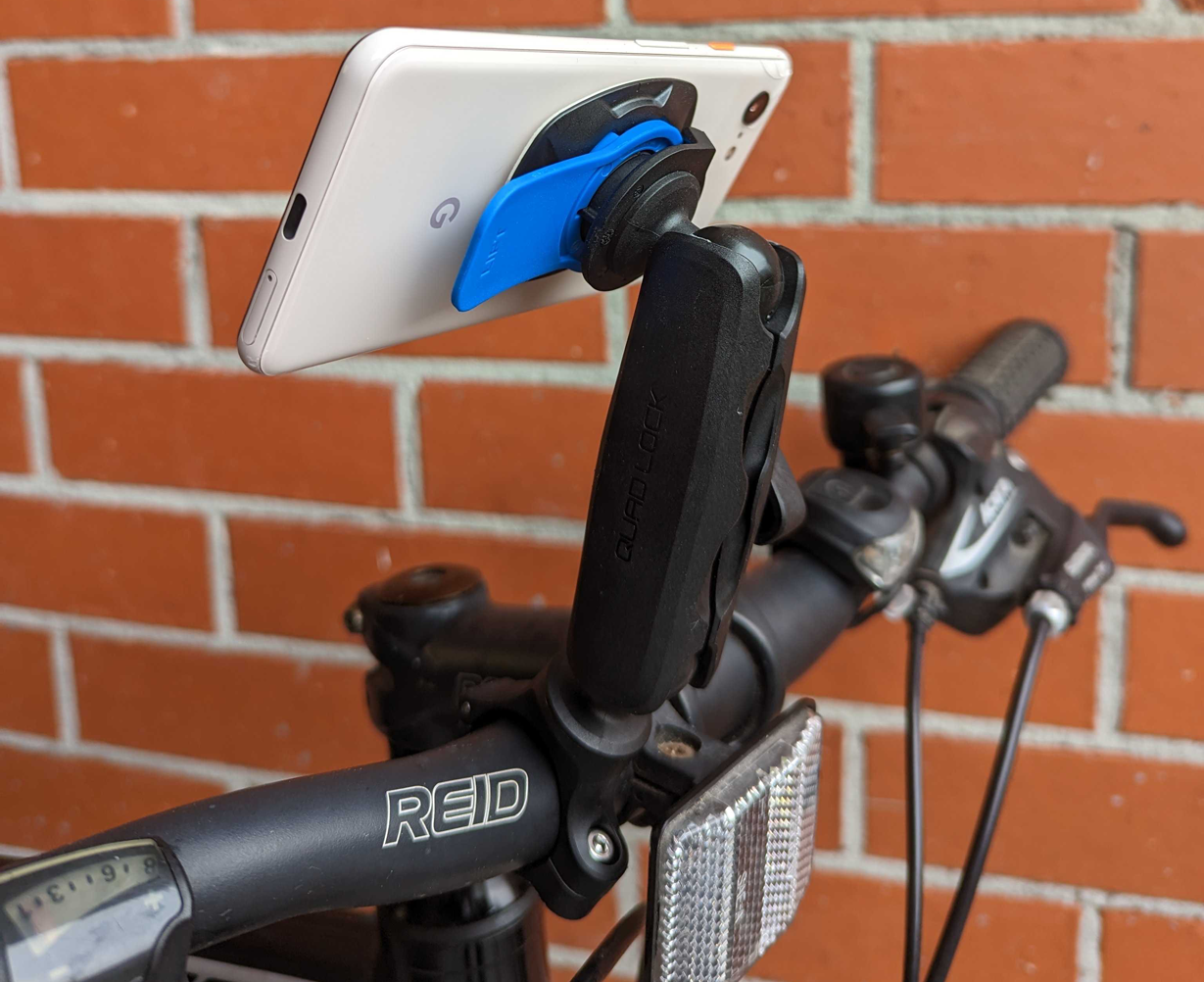}
\caption{
Example of smartphone setup attached to the front handle bars to record onboard street view imagery. 
}
\label{BikeExample}
\end{figure}

Data was gathered by participants who rode their usual cycling routes with a smartphone device attached to the front handlebars. The smartphone device recorded video data at \(1280\text{w} \times 960\text{h}\) resolution at 30 frames per second. Inertial measurement unit (IMU) and global positioning system (GPS) data was time synchronized to image frames and were recorded at 100Hz and 1Hz respectively. An example of the smartphone setup can be seen in Figure.~\ref{BikeExample}.

Data were collected throughout the Greater Melbourne region of Victoria, Australia, over the time period of April to June 2023. Adult participants (aged 18 years and older) were asked to ride as they usually would (and take their usual routes) over a period of approximately one week. The study was approved by the Monash University Human Research Ethics Committee (Project ID: 37431).

A total of 126 participants were recruited, of which 80 participants submitted at least one valid trajectory. Across the 80 participants, a total of 800 valid cycling trips were recorded totaling 360 hours of video footage across 7,006Km of riding.

\section{Video frame labeling pipeline}\label{LabelingSection}  
The aim of the video frame labeling pipeline is to autonomously assign a cycling infrastructure classification label to each of the frames within the video sequence for training the infrastructure classification model in Section. \ref{ModelSection}. Video frame labelling is achieved by first assigning the recorded trajectories’ GPS coordinates into a cycling infrastructure class using the OpenStreetMap \cite{OpenStreetMap} database of known infrastructure. The classified timestamped GPS coordinates are then matched and interpolated to individual video frames resulting in each frame being assigned an infrastructure class.

The labeling pipeline consists of 4 steps:
\begin{enumerate}
\item OpenStreetMap road segment classification
\item GPS coordinate to classified road segment assignment
\item Classified GPS coordinate to video frame assignment
\item Model feedback label correction
\end{enumerate}

The advantage of implementing an automated data labeling pipeline is the savings of both cost and time required to manually label data for training. However automated labelling runs the risk of noisy and imperfect labels being introduced into training data causing inconsistent gradients to be propagated that degrade model performance. Noisy GPS signals, temporary road work construction and incorrect infrastructure labels within the OpenStreetMap database are common errors that can introduce defects into the training data. 
Including model feedback into the labeling pipeline enables abnormality detection of incorrect autonomously generated labels, allowing for targeted manual label correction on a small subset of the data to achieve accurate labels that are time efficient to generate.

\subsubsection{OpenStreetMap road segment classification}
The initial step of the video frame labelling pipeline extracts all road, footpath and cycling path segments from the OpenStreetMap database and assigns them to one of thirteen infrastructure labels. The road segment infrastructure labeling follows \cite{Debjit} to assign roads represented as edges in a graph network to predefined class labels for the Greater Melbourne area.

\subsubsection{GPS coordinate to classified road segment assignment}\label{GPS_road_matching_subsubsection}
\begin{figure*}
\centering
\includegraphics[width=2.0\columnwidth]{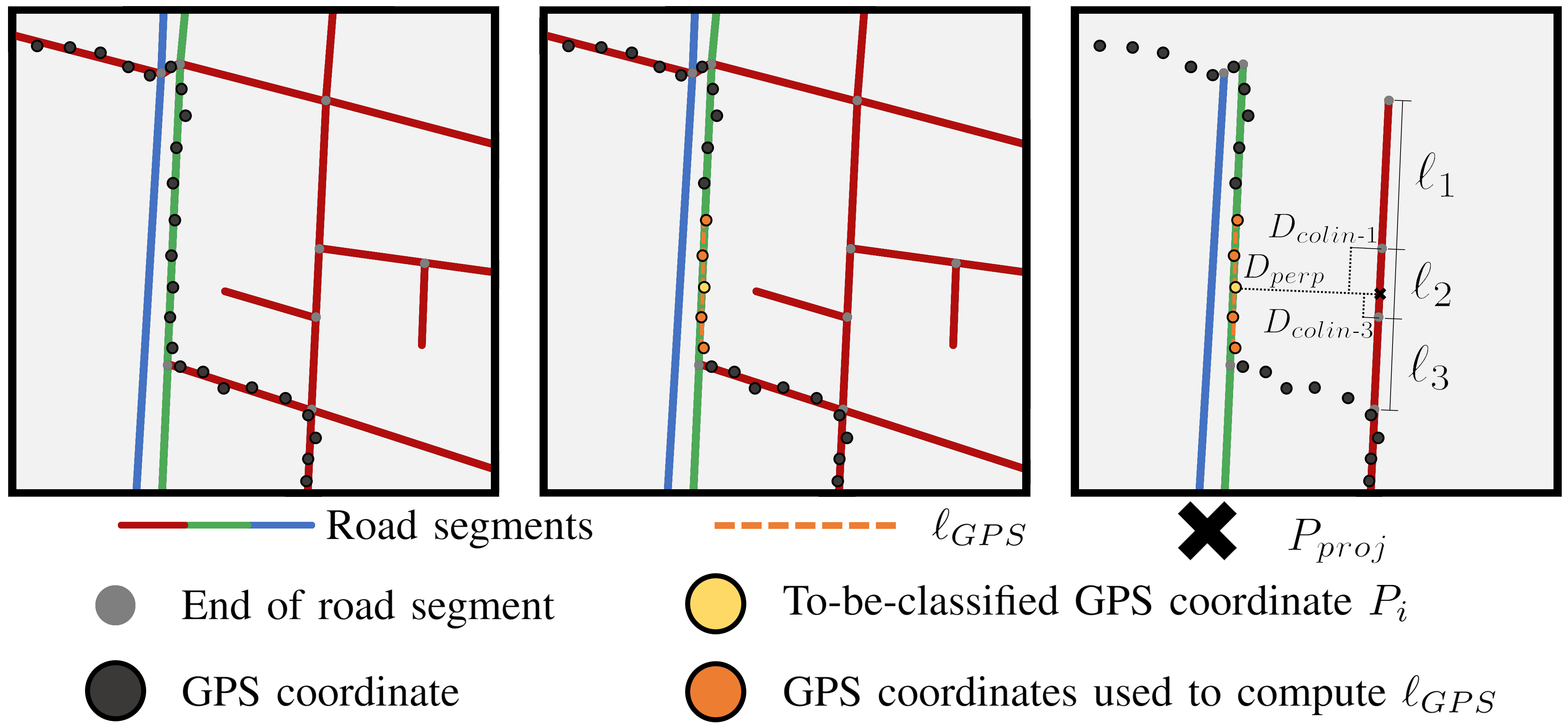}
\caption{
Overview of the GPS to OpenStreetMap road segment assignment algorithm. \\\hspace{\textwidth}
(Left) Example trajectory of cyclist. Each road segment is denoted with a coloured line indicating its respective infrastructure class. The start and end points of the road segment are indicated with a grey circle. \\\hspace{\textwidth}
(Middle) The to-be-classified GPS coordinate \(P_i\) highlighted in yellow samples temporally nearby GPS-coordinates denoted in orange to construct \(\ell_{GPS}\) which is used to filter out non-parallel lines. \\\hspace{\textwidth}
(Right) Non-parallel lines and spatially distant lines are filtered out while remaining lines have their perpendicular and colinear distances (\(D_{perp}\) \& \(D_{colin}\)) computed. For the example shown, \(\ell_1\), \(\ell_2\) \& \(\ell_3\) are colinear to each other, thus share identical perpendicular distances: \(D_{perp\text{-}1} = D_{perp\text{-}2} = D_{perp\text{-}3} \). As the projection of \(P_i\) lies on \(\ell_2\), denoted as \(P_{proj}\), the colinear distance equals zero: \(D_{colin\text{-}2} = 0\). As \(P_{proj}\) lies outside of \(\ell_1\) \& \(\ell_3\) their colinear distances equal \(D_{colin\text{-}1}\) \& \(D_{colin\text{-}3}\) respectively.
}
\label{GPS_matching}
\end{figure*}

To determine which classified road segment belongs to a GPS coordinate, a geometric matching algorithm is used and is summarized in Figure.~\ref{GPS_matching}. The geometric matching algorithm generates a confidence score where lower confidences are given based on the potential for incorrect label assignment due to the presence of multiple adjacent road segments.

For each timestamped GPS coordinate \(P_i\), temporally local GPS coordinates are sampled based on a threshold time distance of the to-be-classified GPS coordinate \(P_i\), for which a 2D line \(\ell_{GPS}\) defined by two end points (\(P_{\ell\text{-}GPS\text{-}1}\) \& \(P_{\ell\text{-}GPS\text{-}2}\)) is fitted using the sampled coordinates. 
A subset of road segment edges \(\ell=\{\ell_1, \ell_2, …., \ell_N\}\) defined as a collection of 2D lines from the Greater Melbourne graph are subsequently proposed given that both the nearest latitude and longitudinal distance from \(P_i\) to line \(\ell_i\) in \(\ell\) and angular distance of \(\ell_{GPS}\) and \(\ell_i\) in \(\ell\) fall below a threshold distance. 

If no lines are proposed, i.e. \(\ell=\{\}\), then the corresponding to-be-classified GPS coordinate \(P_i\) is given a class confidence value of 0.0.
If a single line is proposed i.e. \(\ell=\{\ell_1\}\), or all lines within the set \(\ell\) correspond to identical class labels, then the corresponding to-be-classified GPS coordinate \(P_i\) is assigned the class associated with \(\ell_1\) with a confidence weighting of 1.0, due to no ambiguity in the GPS coordinate’s class assignment. 
If multiple lines  are proposed, i.e. \(\ell=\{\ell_1, \ell_2, …, \ell_N\}\), which comprise of different classes, then a weighting metric based on the perpendicular and colinear distance of \(P_i\) to \(\ell_i\) is used to assign a class and confidence value.

The perpendicular distance of  \(P_i\) to \(\ell_i\) is calculated via: 
\begin{equation}
    D_{perp\text{-}i} = \frac{\ell_i \times (P_{\ell\text{-}i\text{-}1} – P_i)}{\norm{\ell_i}}.
\end{equation}
The colinear distance is calculated by projecting \(P_i\) onto \(\ell_i\) via:
\begin{equation}
    P_{proj\text{-}i} = P_i \pm \frac{D_{Perp-i} 
        \begin{bmatrix}
        0 & -1 \\
        1 & 0 \\
        \end{bmatrix} 
    \ell_i}{\norm{\ell_i}},
\end{equation}
where a check is required to pick the solution that results in \(P_{proj\text{-}i}\) moving towards \(\ell_i\) and not away due to the ambiguity of determining whether a 90 or a 270 degree rotation of \(\ell_i\) is required.
\(P_{proj\text{-}i}\) is then tested to see if it lies between both \(P_{\ell\text{-}i\text{-}1}\) \& \(P_{\ell\text{-}i\text{-}2}\) which results in a colinear distance of 0 (\(D_{colin} = 0\)). If \(P_{proj\text{-}i}\) lies outside of \(\ell_i\), the colinear distance equals the minimum distance between \(P_{proj\text{-}i}\) and \(P_{\ell\text{-}i\text{-}1}\) \& \(P_{\ell\text{-}i\text{-}2}\):
\begin{equation}
    D_{colin\text{-}i} = \text{min}( \norm{P_i - P_{\ell\text{-}i\text{-}1}},  \norm{P_i - P_{\ell\text{-}i\text{-}2}} )
\end{equation}

The final distance for \(P_i\) to \(\ell_i\) is calculated as:
\begin{equation}
    D_i = 1.0 * P_{perp\text{-}i} + 2.0 * D_{colin\text{-}i} + \varepsilon, 
\end{equation}
where \(\varepsilon = 1.0\) to prevent zero-division errors and small distances dominating the final class score. 
The individual distance metrics are then converted to a weighting score by taking the reciprocal and multiplying by the sum of distance metrics 
(\(S_i = \frac{\sum_{j=0}^{N}D_j}{D_i}\)) such that a larger value represents a better fit.
The resultant scores are then sorted into their respective infrastructure classes where the final score for an infrastructure class (\(S_{cls\text{-}j}\)) is the maximum value score that corresponds to it. The infrastructure class with the largest score is assigned to be the infrastructure label for \(P_i\). The confidence in the label is determined by dividing by the sum of all class scores: \(\text{confidence} = \frac{\text{max}(S_{cls})}{\sum_{j=0}^{N}S_{cls\text{-}j}}\).

\subsubsection{Classified GPS coordinate to video frame assignment}
Once each timestamped GPS coordinate along the cycling trajectory \(P_i\) has an infrastructure class associated with it, each timestamped video frame is assigned an infrastructure class and confidence by interpolating between the two temporally closest timestamped GPS coordinates. If a change of infrastructure class occurs between two temporally closest points \(P_i\) \& \(P_{i+1}\) the change in class occurs halfway between the two points and the resultant confidence values for each video frame are halved due to the ambiguity associated with determining when a particular infrastructure starts and stops. 

\subsubsection{Model feedback label correction}
Given a model trained in Section. \ref{ModelSection}, the model is evaluated on the entire dataset where class prediction accuracy metrics for each video trajectory are computed. The accuracy metrics for each video are ranked from worst to best performing where the worst performing videos tended to be resultant from poor label or video quality. 
To ensure consistent gradients being propagated for all infrastructure classes, videos in which the model performed poorly were manually validated to check for label accuracy and video quality. Video trajectories containing poor automatically generated labels were manually corrected. 

The most common errors were associated with incorrect OpenStreetMap infrastructure classification and incorrect GPS coordinate to OpenStreetMap road association. Errors pertaining to incorrect OpenStreetMap road information are resultant from incorrect human labeling of cycling infrastructure and newly built or temporary (road works) infrastructure not being updated in the database. For errors related to incorrect GPS coordinate to OpenStreetMap road assignment as described in Section \ref{GPS_road_matching_subsubsection}, the most common error was caused due to the degradation of GPS quality in metropolitan areas and the higher density of road segments associated with these areas (such as having an off-road path located directly next to a road). Videos which were deemed of poor quality were removed from the training data, which was typically due to the camera being incorrectly mounted.

\section{Results}
\subsection{Model performance}\label{ModelPerformanceSection}
The proposed model was trained on a training dataset consisting of 1.1 million images, 974k (88.00\%) of those with a valid infrastructure label weighting above zero and 232k (20.98\%) which were manually validated. 
The model was then validated on a withheld dataset consisting of 14.3k images, 12.9k (90.30\%) of which were manually validated. Only the 12.9k images that were manually validated were used for the computation of evaluation metrics.

\begin{figure}[b]
\centering
\includegraphics[width=1.0\columnwidth]{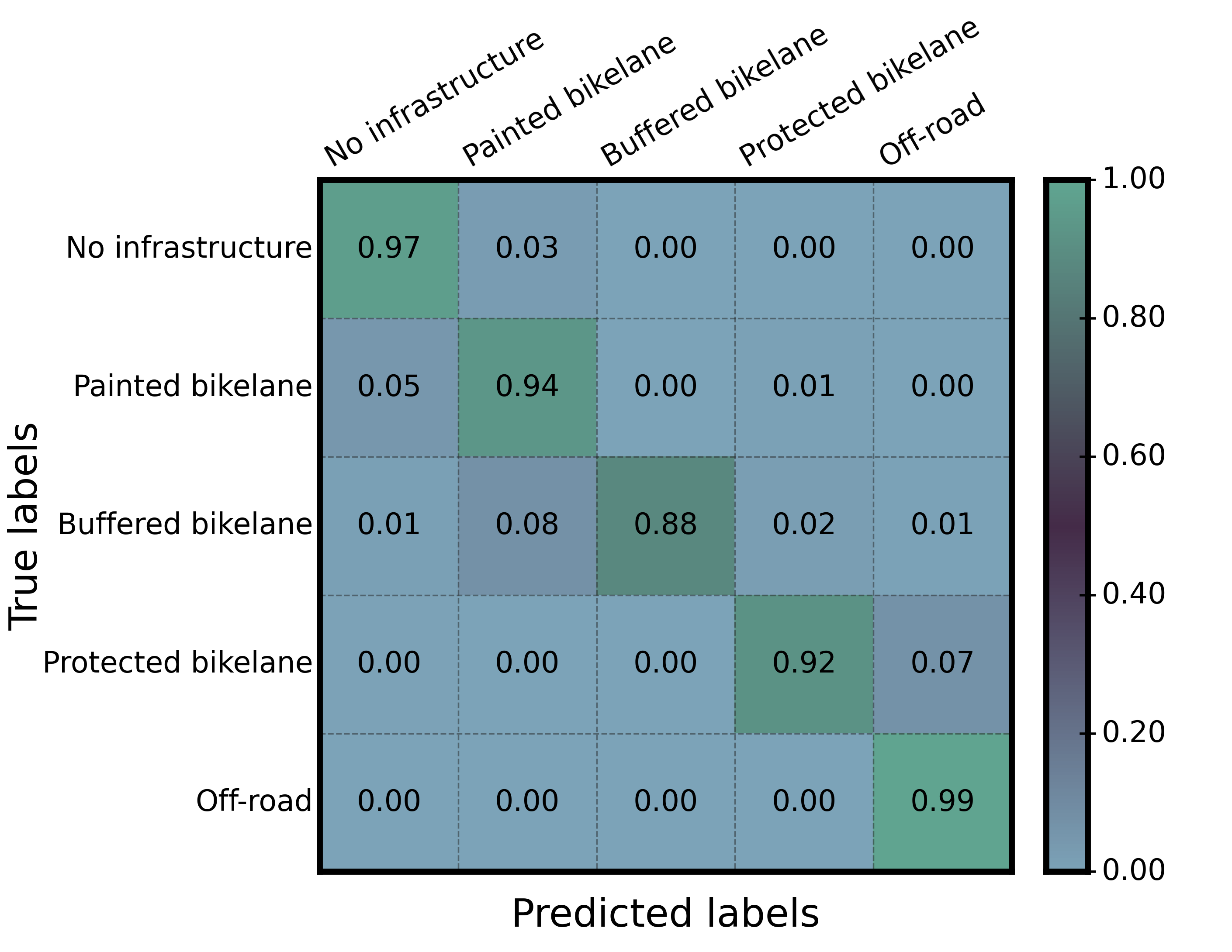}
\caption{Normalized confusion matrix of the proposed model’s main classes for the validation dataset.}
\label{ConfusionMatrixFig}
\end{figure}

The results of the proposed model can be seen in Figure.~\ref{ConfusionMatrixFig}. The model achieved an average main class accuracy result of 96.25\% where all classes were able to attain an accuracy of over 90\% with the exception of the buffered bike lane infrastructure. The lower accuracy of the buffered bike lane infrastructure class was due to the shared visual features between it and the painted bike lane class infrastructure, causing an 8\% confusion in predictions.

\subsection{Ablation study}\label{AblationSect}
To assess model performance and the utility of temporal information for solving the task, three models were evaluated in an ablation study: 
\begin{enumerate} 
\item \textbf{Proposed model}: The full model as defined within section~\ref{ModelSection}.
\item \textbf{No future}: The model only has access to current and prior images of a sequence when predicting infrastructure classes of the current timestep. The attention matrix is masked during training and evaluation to only include current and prior information. The model aims to measure the reliance on both past and future information.
\item \textbf{No temporal}: The no temporal model omits the temporal extractor network component and only uses backbone training as per section~\ref{BackBoneTrainingSect}. The aim of the no temporal model is to assess the utility of temporal information in predicting infrastructure class labels.
\end{enumerate}

Each of the aforementioned three models were trained and then validated using the datasets as specified in section~\ref{ModelPerformanceSection}. Each model was trained over 3 random initializations.
Both the proposed and no future models were given a sequence of 50 images as input. For each input sequence of 50 images, a total of 10 output classifications were recorded. For the proposed model, the 10 recorded output classifications were sampled from the center of the image sequence, while the no future model sampled from the 10 most recent images. The next sequence of 50 images were sampled using a stride length of 10, where sampled image indices that fall outside of the image sequence range are replaced with black images. For the no temporal model, a single image was given as input to produce a single output classification.

The results of the ablation study can be found in Table~\ref{ablationTable}.
The proposed model achieved a main and sub-class accuracy of 96.25\% and 95.38\% respectively on the validation dataset. The use of future information accounted for an increase in sub-class accuracy of 4.25\% whilst the use of temporal information resulted in an accuracy increase of 7.55\%. The results demonstrate that estimating cycling infrastructure from onboard imagery is dependent on temporal information from both past and future. 

\begin{table}[b]
    \centering
    \caption{}
    \begin{tabular}{|l|c|c|}
        \hline
        & Main class accuracy & Sub class accuracy\\
        \hline
        Proposed model & 96.25\% (6.60) & 95.38\% (33.31) \\
        \hline
        No future & 92.86\% (8.49) & 91.13\% (36.90) \\
        \hline
        No temporal & 90.38\% (9.09) & 87.83\% (35.15) \\
        \hline
    \end{tabular}
    Ablation study results. Values within parentheses indicate standard deviations of the metric calculated using each initialization and each individual class accuracy measurement.
    \label{ablationTable}
\end{table}

\subsection{Image blackout study}
\begin{figure}
\centering
\includegraphics[width=0.98\columnwidth]{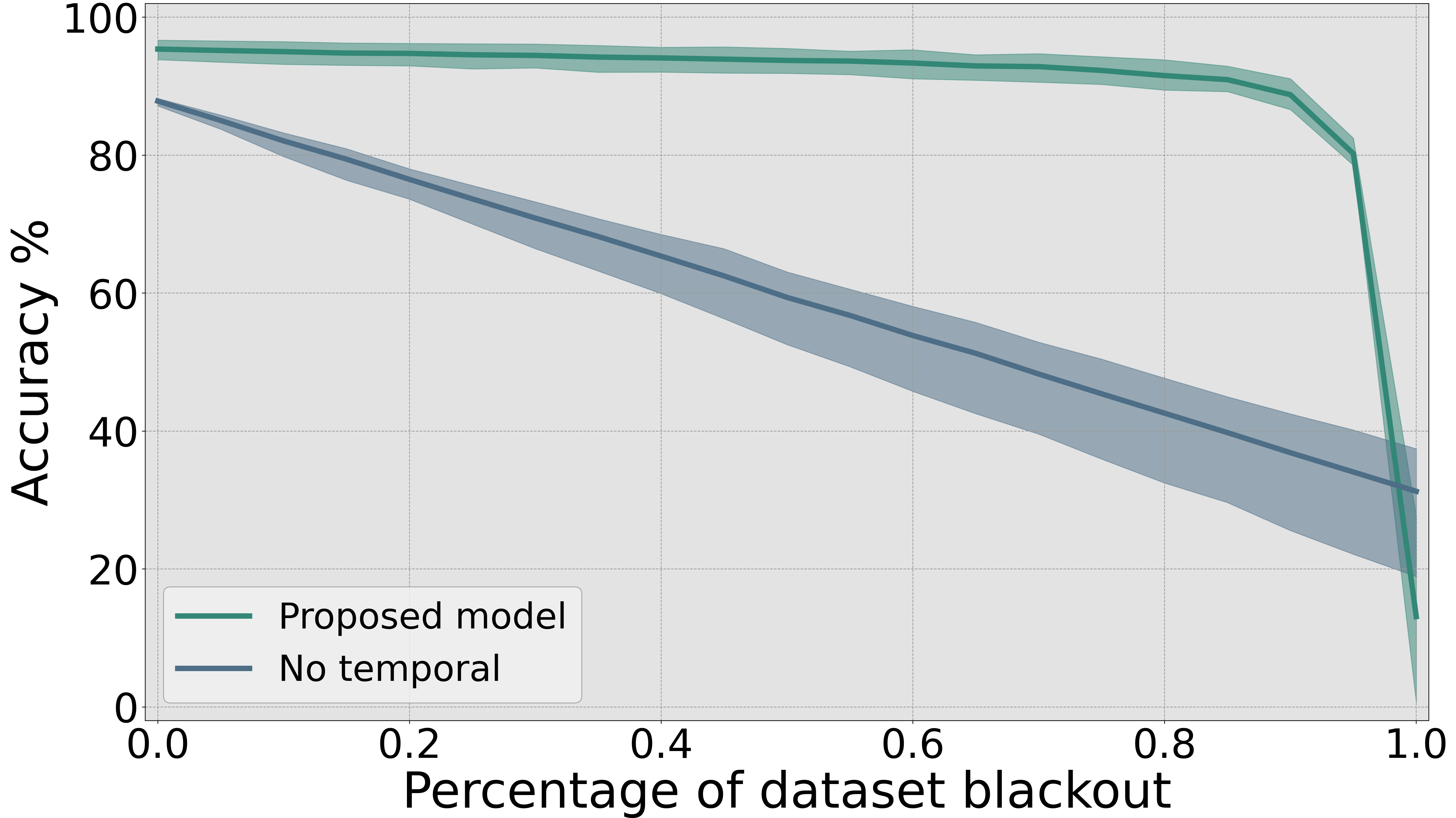}
\caption{
Results of image blackout study. Thick lines indicate the average sub-class accuracy over the six trials whilst the shaded regions denote the minimum and maximum accuracy.
}
\label{BlackoutFig}
\end{figure}

To measure the model’s robustness to correctly estimate cycling infrastructure from images that contain no identifying class features, i.e. the absence of visible signage, an experiment was held where datapoints were replaced with black images. Both the proposed model and the no temporal model were tested on the validation dataset outlined in section~\ref{AblationSect}, where images within the dataset were probabilistically replaced with black images. Each model was passed through the dataset where blackout probabilities from 0.0 to 1.0 using 0.05 increments were used. To account for blackout sampling variation, each model and their associated 3 training initializations were passed through the dataset a total of 2 times, resulting in a total of 6 accuracy datapoints per blackout probability. Images that are blacked out are still required to have the cycling infrastructure label associated with them estimated and are included in the accuracy computation.  The results of the feature absence study can be found in Figure.~\ref{BlackoutFig}. 

The results demonstrate the importance of the temporal component of the model for maintaining robustness when information is only intermittently available. The proposed model maintained its ability to correctly classify cycling infrastructure even when most frames were blacked-out. A performance drop of 6.61\% in the sub-class accuracy was recorded when transitioning from blackout probabilities 0\% to 90\%. After which rapid degradation in model performance was recorded where sub-class accuracy fell a further 8.54\% when transitioning from blackout probability 90\% to 95\%. 
Compared to the no temporal model which observed a consistent linear decrease in performance, the proposed model was able to match the no temporal model’s base accuracy of 87.83\% (blackout probability 0\%) at a blackout probability of 90\% with a sub-class accuracy of 88.77\%. 

\section{Discussion}
The main limitation of the proposed approach are the inaccuracies induced within the autonomous data labeling pipeline due to OpenStreetMap and GPS related errors, requiring laborious manual label correction. Although model feedback allowed for targeted error correction, a major error experienced was the incorrect labeling of a specific type of protected bike lane within the OpenStreetMap database as a dedicated off-road bike path. This error caused the model to consistently classify certain protected bike lanes as dedicated off-road bike paths which were undetectable via model feedback due to both sharing identical incorrect labels. The solution required the manual correction of trajectories containing large portions of dedicated off-road bike paths until model feedback correction overcame the biases induced by OpenStreetMap errors.

Labeling errors associated with poor GPS quality around metropolitan areas due to poor signal strength coupled and high density of road segments from such areas was another prominent source of errors. Although such errors could potentially be solved using path connectivity analysis of the OpenStreetMap graph network, it is observed that cyclists often switch between infrastructure outside of the OpenStreetMap defined node to edge transitions due to not capturing infrastructure transition points such as curb cuts and driveways.

Another limitation is the local specificity of training data. Although the data covers a diverse range of infrastructure and environmental conditions, the generality of the model to transfer to unseen infrastructure standards outside of Australia and weather conditions such as snow are unknown.

\section{Conclusion}
In this work we propose an approach to estimate cycling infrastructure using onboard street-level imagery. The approach comprised of four components: a backbone to extract image features, an encoder to compress features onto a latent vector, a temporal extractor to analyse temporal information and a decoder to estimate infrastructure classes. The model receives an image sequence as input and outputs cycling infrastructure class labels based on a hierarchical classification system.

Data was generated from 360 hours of participant gathered cycling, covering 7,006Km of riding across the greater Melbourne area. Data was automatically labeled via a video frame labeling pipeline which utilizes onboard GPS information and the OpenStreetMap database of known infrastructure.

The model achieved a main and sub class accuracy of 96.25\% and 95.38\% respectively, attaining a sub class accuracy improvement of 7.55\% compared to when omitting temporal information. The proposed model demonstrated robustness to the complete absence of image features where a minor degradation of 6.61\% in classification accuracy was observed after 90\% of data was replaced with blank images.

Further work is required to identify additional features such as parked cars (temporary features) and lane widths (static features). A further understanding of daily trends of temporary features can be attained by extending the approach to crowd-sourced data to create dynamic infrastructure maps. Collectively, these approaches can then be used to identify gaps in connectivity of the bicycle network, and identify where reallocation of road space for bicycle infrastructure is feasible (e.g. the removal of car parking spaces for the provision of protected bicycle facilities). These capabilities can then be leveraged to identify infrastructure improvements that offer the greatest cost-benefit investment to improve safe cycling.

\bibliographystyle{IEEEtran}
\bibliography{bibliography}

\end{document}